# A Cooperative Positioning Flamework for Robot and Smart Phone Based on Visible Light Communication

**Junye Chen[1], Fangdi Li[1], Futong An[2], Chen Yang[1], Hongzhan Song[1], Shangsheng Wen[3, *], Weipeng Guan[3]**

[1]School of Automation Science and Engineering, South China University of Technology, Guangzhou, Guangdong 510640, China
[2]School of Information Engineering, South China University of Technology, Guangzhou, Guangdong 510640, China
[3]School of Materials Science and Engineering, South China University of Technology, Guangzhou, Guangdong 510640, China
* Correspondence: shshwen@scut.edu.cn

**Abstract:** A cooperative positioning flamework of human and robots based on visible light communication (VLC) is proposed. Based on the experiment system, we demonstrated it is feasible and has high-accuracy and real-time performance.

## 1. Overview

Benefiting from the popularity of light-emitting diode (LED) lighting technology and the rapid development of high-resolution complementary metal-oxide-semiconductor (CMOS) sensor in recent years, indoor positioning technology based on image sensors has ushered in vigorous development and broad prospects. Because the walls block the transmission of signals, the performance of traditional positioning technologies such as Global Positioning System (GPS) positioning decreases significantly when used indoors, which has a large positioning error that even could not be used at all. Visible light positioning (VLP) systems compile the identifier (ID)-position information into a modulated signal by Microcontroller Unit (MCU) and adopt on-off keying (OOK) modulation to modulate the LEDs, transmitting information by changing the on-off state of the LEDs. Then, for the receiving terminal, by the rolling shutter effect of CMOS, the on-off state of the LED is captured in the form of stripes, so that the data rate of optical camera communication (OCC) can be boosted to a level much higher than the video frame rate. Hence, high-speed VLC communication can be established. Next, each unique identifier (UID) assigned to an LED corresponds to an actual position and the relation is saved in the map database. In this case, the receiving terminal can obtain the actual position of an LED by decoding the stripes, and further determining its position on the map. So far, some proposed positioning methods based on smartphones have achieved high accuracy in experiments. In addition, in [1], a positioning method for robots based on the robot operating system was once proposed. In [2, 3], a robot positioning method based on a single LED was proposed, and combined with simultaneous localization and mapping (SLAM), which can reach a high accuracy of 2.5 cm and achieve navigation.

In warehouse transportation, industrial production, commercial service, and other scenarios, the cooperation of humans and robots is in great demand and becomes available. In dispatching command and navigation, the cooperative location of robots plays an important role and is demonstrated to be feasible [4]. Moreover, the position of the controllers or the objects that the robots serve is changing from moment to moment in practical application, which requires not only being able to master the trajectory of the robots in real-time but also to track the position of human with a smartphone at all times. Moreover, they should be capable of sharing locations with each other. Therefore, it is challenging to cooperative position mobile devices and robots in a complex and unpredictable scenario and the work in this field is meaningful and significant.

## 2. Innovation

In this demonstration, we propose a cooperative positioning framework for smartphones and robots. The accuracy of the ID identification, positioning accuracy, and real-time are mainly studied and the effectiveness of this scheme is experimentally verified. The main contributions of this demonstration are as follows:

1. Design a VLC high-accuracy cooperative positioning system, which is applied to cooperative positioning on smartphones and robots. Since VLP on smartphones needs to consider the tilt posture in practical application, this system applies several different VLP schemes to cope with different lighting situations [5-7].

2. Based on the methods mentioned above, we designed and built a cooperative positioning framework for smartphones and robots. The locations of the smartphones and robots can be obtained on the smartphone in real-time.

## 3. Description of Demonstration

The demonstration of our positioning system consists of two parts: the modulated LED transmitters and the position receive terminals. The overall experimental environment and result for our demonstration are shown in Fig. 1. Four LED transmitters installed on the flat plates are applied to broadcast the actual position information. In order to easy design and be scalable, the control circuit unit is made up of several off-the-shelf modules. As is shown in Fig. 1, the AC to DC power supply converts AC220V to DC36-72V, providing power supply for the LED driver module and MCU buck module. Then, the MCU driven by the MCU buck module generates the control signal to control the amplification module to drive the LED transmitter. In addition, the two buttons beside the MCU can change the UID and frequency of the transmitter respectively. In the demonstration, the default UID information of the lamp is downloaded in the memory of the MCU in advance, converted into high and low levels automatically, and generates the control signal to modulate the light of those LEDs by adopting the OOK modulation scheme, which achieves the signal transmission. The LEDs we utilized with a diameter of 175mm or a side of 175mm, are mounted on poles of flat plates with a height of 2.5m.

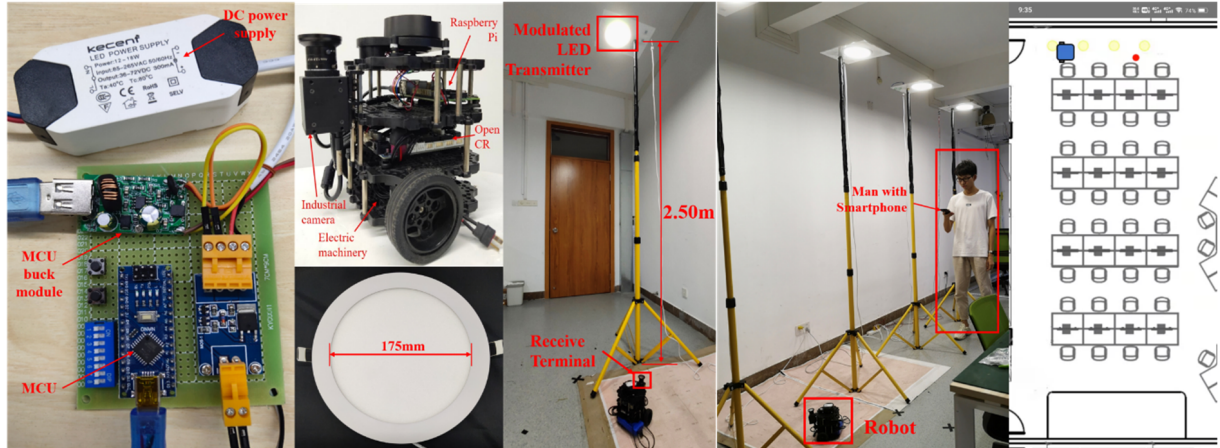

Fig.1. Demonstration devices and experimental environment of cooperative positioning flamework based on visible light communication.

The other part is the receive terminals of signal receiving and positioning. To realize positioning in different terminals, the first difficulty is that different receiver cameras and lenses have different parameters. We proposed an adaptive positioning method that is independent of the hardware parameters. When the receiver enters the illumination area, the LEDs are captured in an image by the camera simultaneously. By image processing technology, the ROI can be extracted from the image with LED-ID detected method, based on which then the UID is identified by the decoding method [8]. Furthermore, referring to the UID information in the pre-established UID location database, the actual location of the LEDs can be obtained. Meanwhile, after image processing, the contours of LEDs extracted in ROI are applied to the VLC positioning algorithm. Thus, the relative position of the terminal refers to the reference LED (specified according to the UID) can be obtained and combining the actual location of the reference LED mentioned above can finally obtain the actual position of the terminal. The location of robot can be obtained by the robot positioning method and transmitted to the smartphone [9]. So the real-time locations of the human and robot can be easily tracked in the app on the smartphone in Fig. 1 and further realize the robot navigation. The red point is the positioning results of the human (smartphone), the blue block with two wheels is the positioning results of the robot with a camera and the yellow rounds are the LED transmitters. The MCU of the control unit is ATMELATMEGA328P-AU. The operating system of the Turtlebot3 robot is Ubuntu 18.04 MATE. And the system of the remote control is Ubuntu 18.04 desktop. The system uses the Melodic version of ROS.

4. OFC Relevance

The implementation deals high-accuracy positioning system based on VLC that is popular in the indoor localization service. By deploying the system on the human-robot cooperative positioning platform, real-time human-robot interaction can be realized, which has engaging prospects in manufacturing and commercial. For OFC being the largest-scale conference on optical networks and communication, the completely proposed design aims to spark a widespread novel application and research of VLC.

*Acknowledgement*
This research was funded by Research and Development Program in Key Areas of Guangdong Province (2019B010116002); National Undergraduate Innovation and Entrepreneurship Training Program (202010561020, 202110561163); Guangdong Science and Technology Project under Grant (2017B010114001).

Demonstration video is available at:
https://www.bilibili.com/video/av686451729/?zw&vd_source=0397df10121178e96480661d23ed58e7
https://www.bilibili.com/video/av302876024/?zw&vd_source=0397df10121178e96480661d23ed58e7
https://www.bilibili.com/video/av636014353/?zw&vd_source=0397df10121178e96480661d23ed58e7